\documentclass[conference]{IEEEtran}
\IEEEoverridecommandlockouts

\usepackage{cite}
\usepackage{amsmath,amssymb,amsfonts}
\usepackage{algorithmic}
\usepackage{graphicx}
\usepackage{textcomp}
\usepackage{xcolor}
\usepackage{multirow}
\usepackage{booktabs,makecell,siunitx}
\usepackage{hyperref}

\def\BibTeX{{\rm B\kern-.05em{\sc i\kern-.025em b}\kern-.08em
    T\kern-.1667em\lower.7ex\hbox{E}\kern-.125emX}}
\begin{document}

\title{UltraFlwr - An Efficient Federated Surgical Object Detection Framework\\
\thanks{This work was supported by King’s-China Scholarship Council PhD Scholarship programme (K-CSC).}
}

\author{
\IEEEauthorblockN{
Yang Li\IEEEauthorrefmark{1}\textsuperscript{\P\#},
Soumya Snigdha Kundu\IEEEauthorrefmark{1}\textsuperscript{\#},
Maxence Boels\IEEEauthorrefmark{1},
Toktam Mahmoodi\IEEEauthorrefmark{4},
}
\IEEEauthorblockN{
Sebastien Ourselin\IEEEauthorrefmark{1},
Tom Vercauteren\IEEEauthorrefmark{1},
Prokar Dasgupta\IEEEauthorrefmark{1}\IEEEauthorrefmark{3},
Jonathan Shapey\IEEEauthorrefmark{1}\IEEEauthorrefmark{2},
Alejandro Granados\IEEEauthorrefmark{1}
}
\IEEEauthorblockA{\IEEEauthorrefmark{1}Surgical \& Interventional Engineering, King's College London, UK}
\IEEEauthorblockA{\IEEEauthorrefmark{2}Neurosurgery Department, King's College Hospital, UK}
\IEEEauthorblockA{\IEEEauthorrefmark{3}Department of Urology, Guy's Hospital, UK}
\IEEEauthorblockA{\IEEEauthorrefmark{4}Department of Engineering, King's College London, UK}
\IEEEauthorblockA{\textsuperscript{\P}Corresponding author: yang.7.li@kcl.ac.uk}
\IEEEauthorblockA{\textsuperscript{\#}These authors contributed equally to this work.}
}

\maketitle

\begin{abstract}
Surgical object detection in laparoscopic videos enables real-time instrument identification for workflow analysis and skills assessment, but training robust models such as You Only Look Once (YOLO) is challenged by limited data, privacy constraints, and inter-institutional variability.
Federated learning (FL) enables collaborative training without sharing raw data, yet practical support for modern YOLO pipelines under heterogeneous surgical data remains limited. 
We present \emph{UltraFlwr}, an open-source, communication-efficient, and edge-deployable framework that integrates Ultralytics YOLO with the Flower FL platform and supports native Partial Aggregation (PA) of YOLO components (backbone, neck, head). 
Using two public laparoscopic surgical tool detection datasets, we conduct a systematic empirical study of federated YOLO training under Independent and Identically Distributed (IID) and multiple clinically motivated heterogeneous scenarios, including differences in data curation, video length, and label availability. 
Results show that standard FL aggregators (e.g., FedAvg) do not consistently match centralized training per client, but reduce inter-client performance variability. 
Aggregating both backbone and neck components achieves performance comparable to full aggregation with lower communication costs. 
Also, improving within-client data consistency can benefit FL even when it increases distribution shift across clients. 
These findings provide practical guidance for deploying federated YOLO-based object detection in heterogeneous surgical environments.
\end{abstract}


\section{Introduction} \label{sec:intro}
Surgical object detection in laparoscopic videos is a foundational capability for surgical workflow analysis and skill assessment, enabling real-time identification and tracking of surgical instruments \cite{Pan2024DBHYOLO}. 
Among existing approaches, the You Only Look Once (YOLO) family of models has consistently achieved State-Of-The-Art (SOTA) performance for real-time object detection, offering a favorable trade-off between accuracy and inference speed \cite{Jocher_2023_Ultralytics_YOLO}. 
Despite these advances, developing and deploying YOLO-based models in surgical settings remains challenging. 
Surgical procedures exhibit substantial variability across cases, surgeons, and institutions \cite{sevik2024surgtechniquereview}, requiring large and diverse datasets for robust model generalization. 
However, annotated surgical data are scarce due to ethical constraints, high annotation costs, and the need for domain expertise \cite{Mohammed2024YOLOReview}. 
In addition, regulatory frameworks such as the Health Insurance Portability and Accountability Act (HIPAA) and the General Data Protection Regulation (GDPR) impose strict limitations on cross-institutional data sharing \cite{Amitav2025GDPR}. 
Even in the absence of regulatory barriers, the sheer volume of video data generated by surgical procedures renders centralized storage and processing across institutions impractical.

Federated Learning (FL) offers a principled solution to these challenges by enabling collaborative model training across institutions without sharing raw data \cite{mcmahan2017FL}. 
In FL, models are trained locally at each client site, and only model updates are communicated to a central server for aggregation. 
This aggregation step is central to FL, allowing knowledge to be shared while preserving data privacy and institutional autonomy. 

However, applying FL to surgical object detection introduces additional challenges. 
The variability inherent to surgical practice induces statistical heterogeneity across clients, where data are independently sampled from client-specific distributions rather than a shared global distribution. 
Beyond procedural differences, heterogeneity arises from variations in data curation, annotation conventions, and imaging hardware. 
In laparoscopic video datasets, this manifests as shifts in sample size, label distribution, annotation density, video length, frame sampling frequency, and bounding box statistics. 
Such heterogeneous data distributions can lead to client drift, adversely affecting convergence and generalization in FL \cite{Lu2024FLNonIIDSurvey}.

In parallel, large-scale clinical deployment of FL systems is constrained by communication efficiency and environmental considerations. 
Frequent exchange of large model updates incurs substantial bandwidth consumption, latency \cite{shahid2021communicationefficiencyFL}, and carbon emissions \cite{Thakur2025GreenFL}. 
These costs are particularly pronounced in surgical settings, where data are predominantly video-based and collaboration typically occurs at the hospital-server level rather than over geographically distributed infrastructures. This shifts model training closer to the data source, necessitating edge computing–based FL that limits unnecessary data movement.

To address these challenges, we introduce \emph{UltraFlwr}, a communication-efficient and edge-deployable FL framework for federated YOLO training in networked surgical environments. 
UltraFlwr integrates the Ultralytics YOLO implementation \cite{Jocher_2023_Ultralytics_YOLO} with the Flower FL platform \cite{beutel2022flower}, enabling flexible and scalable federated training of YOLO models. 
The framework supports Partial Aggregation (PA) strategies that selectively aggregate model components to reduce communication overhead while maintaining, and in some cases improving, detection performance compared to Full Aggregation (FA) using standard FL aggregators such as FedAvg \cite{mcmahan2017FL} and FedMedian \cite{Yin2018FedMedian}. 
UltraFlwr is open-sourced at \url{https://github.com/KCL-BMEIS/UltraFlwr}.

Using UltraFlwr, we conduct a systematic empirical evaluation of federated YOLO training under both Independent and Identically Distributed (IID) and clinically motivated heterogeneous settings. 
These scenarios capture heterogeneity arising from differences in data curation, video length, and label availability, providing practical insights into federated YOLO deployment in surgical environments. 
Specifically, we investigate the following Research Questions (RQs):

\begin{itemize}
    \item \textbf{RQ1:} Is FL non-inferior to single-institution centralized training in terms of detection performance, while enabling privacy-preserving multi-institution collaboration?
    \item \textbf{RQ2:} How do different PA strategies affect the trade-off between communication efficiency and detection performance in federated YOLO training?
    \item \textbf{RQ3:} How do different forms of heterogeneity commonly observed in surgical datasets influence YOLO performance under FL?
\end{itemize}

\section{Related Work}
\subsection{FL for Medical Object Detection}

Prior work on federated medical object detection has primarily focused on radiological imaging, most notably through the federated extension of nnDetection \cite{Rashidi2024FednnDetection}.
The authors show that federated training across simulated institutions achieves performance comparable to centralized training while allowing to preserve data privacy. 
Their analysis highlights the impact of distributional heterogeneity, including label-based heterogeneous settings that induce bias toward majority classes, as well as manufacturer-based heterogeneity arising from differences in imaging devices. 
Notably, manufacturer-based heterogeneity was found to degrade performance more severely than label imbalance in radiological data. 
In contrast, the impact of label imbalance and other clinically relevant heterogeneities under FL remains largely unexplored for surgical object detection.


\subsection{Partial Model Aggregation in FL}
 PA in FL is introduced to address data heterogeneity by allowing clients to retain personalized model components while sharing a subset of parameters \cite{arivazhagan2019fedper}. 
 Subsequent works extended this paradigm through more flexible formulations, including decoupling global and local representations or selectively updating shared components \cite{wang2024partial,oh2022fedbabu}, as well as system-level co-optimization of communication and computation resources \cite{Chen2024EfficientFL}. 
 However, PA has been studied predominantly in other settings and remains largely unexplored for object detection. 
 In particular, the impact of partially aggregating modular detector components (e.g., backbone, neck, head) on detection performance, convergence, and communication efficiency in FL is not yet well understood.

\begin{table}[t]
\centering
\caption{Parameters saved per client during federated training for different YOLO-PA strategies with YOLOv11n, compared to standard FA. Values represent the reduction in parameter communication costs per client per round. The fewer parameters transmitted, the better.}
\label{tab:parameter_savings}
\begin{tabular}{c c}
\toprule
\textbf{Strategy} & \textbf{Params. Saved (\# and \%)} \\
\midrule
Full Aggregation & 0 (0\%) \\
FedBackbone & $1,034,061$ (39.17\%) \\
FedNeck & $2,073,798$ (78.56\%) \\
FedHead & $2,171,791$ (82.27\%) \\
FedNeckHead & $1,605,764$ (60.83\%) \\
FedBackboneHead & $566,027$ (21.44\%) \\
FedBackboneNeck & $468,034$ (17.73\%) \\
\bottomrule
\end{tabular}
\end{table}

\section{Methods}
\subsection{UltraFlwr Framework}

UltraFlwr integrates the Ultralytics YOLO implementation with the Flower FL platform \cite{beutel2022flower}, enabling scalable and configurable federated training of YOLO models. 
Combining strengths from Ultralytics and Flower, UltraFlwr supports standard FL configurations, including client participation, communication rounds, and aggregation strategies, and natively implements PA of YOLO components (backbone, neck, and head) to reduce communication overhead per FL round. 
In addition, UltraFlwr provides built-in utilities for training monitoring and performance evaluation.

UltraFlwr supports multiple YOLO variants to accommodate heterogeneous resource constraints. 
In this work, we adopt YOLOv11n, a lightweight model designed for edge deployment, to emphasize communication efficiency and practical feasibility in federated surgical settings.

\subsection{YOLO-PA: Native YOLO-Based Partial Aggregation Strategies}

YOLO models follow a modular architecture comprising a backbone, neck, and head \cite{Jocher_2023_Ultralytics_YOLO}. 
The backbone extracts visual features using a Cross Stage Partial (CSP)-style Convolutional Neural Network (CNN) \cite{Terven2023YOLOReview}. 
The neck aggregates multi-scale features, typically via Feature Pyramid Network (FPN) and Path Aggregation Network (PAN) designs \cite{lin2017feature,Terven2023YOLOReview}. 
The head performs object detection by predicting bounding boxes and class probabilities. 
UltraFlwr exploits this modularity to implement native PA strategies that selectively aggregate YOLO components during federated training.

We consider PA strategies grouped by the number of aggregated components. 
Single-component aggregation updates only one module per round: Backbone (FedBackbone), Neck (FedNeck), or Head (FedHead). 
Dual-component aggregation updates pairs of modules: Backbone-Head (FedBackboneHead), Backbone-Neck (FedBackboneNeck), and Neck-Head (FedNeckHead). 
All strategies are compatible with well-used aggregation rules, including FedAvg \cite{mcmahan2017FL} and FedMedian \cite{Yin2018FedMedian}.

YOLO-PA substantially reduces the number of parameters transmitted and aggregated per communication round without modifying the underlying model architecture. 
For YOLOv11n, aggregating only the detection head reduces communication costs by 82.27\% relative to FA (Table~\ref{tab:parameter_savings}). 
This absolute parameter reduction increases for larger YOLO variants, as their backbone and neck contain a greater number of parameters, leading to larger raw communication savings under PA.

\subsection{Datasets}
\begin{figure}[t]
    \centering
    \includegraphics[width=\columnwidth]{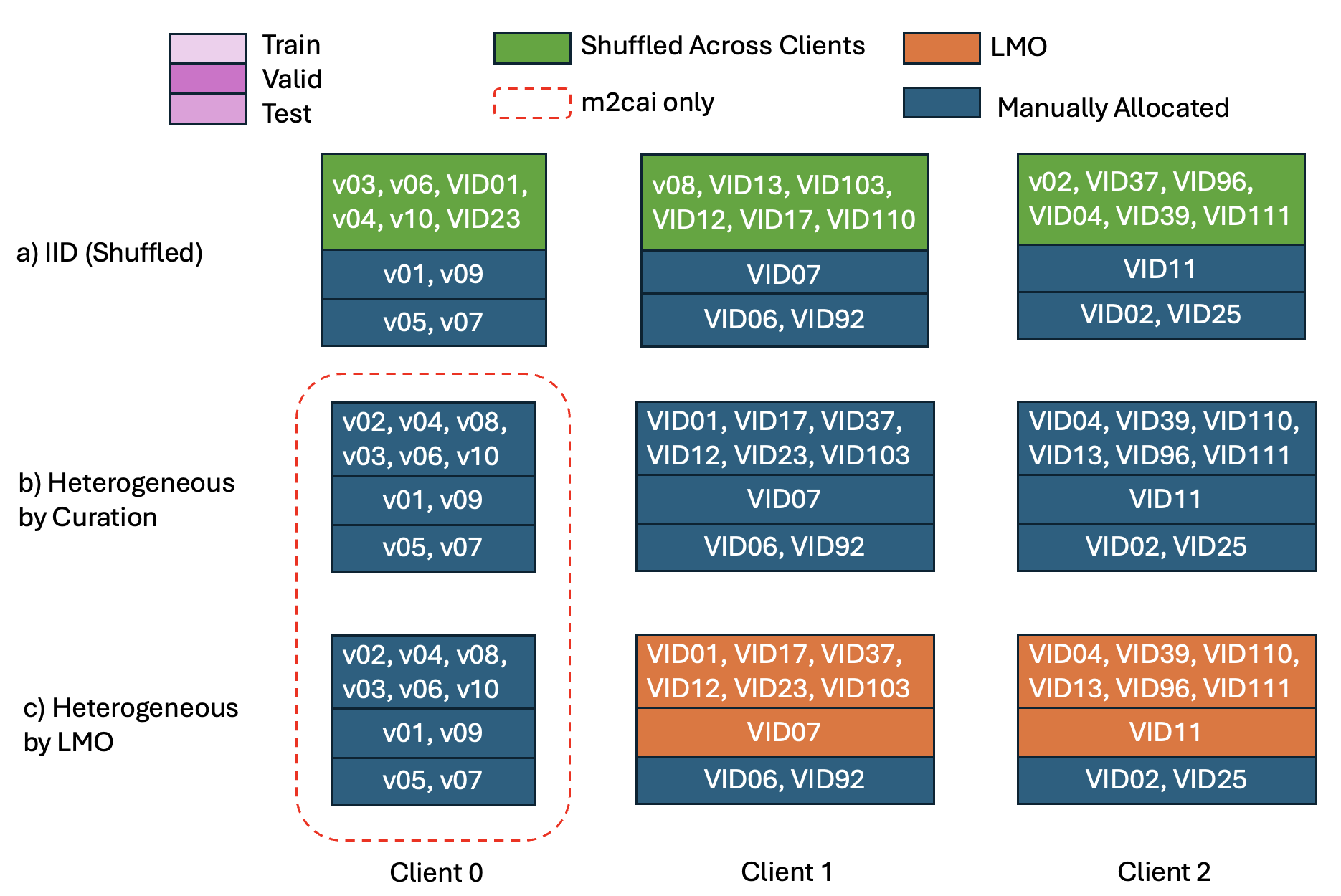}
    \caption{Dataset distribution across clients in the $G_{\text{combined}}$ experimental settings, constructed from m2cai16-tool-locations and CholecTrack20. 
    Each client box shows train/validation/test splits. 
    Three settings are considered: (a) IID ($G_{\text{combined-a}}$), where training videos are pooled and randomly redistributed across clients (green); 
    (b) curation-based heterogeneity ($G_{\text{combined-b}}$), where client 0 contains only m2cai16-tool-locations videos (v*), while clients 1 and 2 contain disjoint subsets of CholecTrack20 videos (VID*); 
    and (c) Leave-Multiple-Out (LMO) label heterogeneity ($G_{\text{combined-c}}$), where client 0 retains annotations for all tool classes, while clients 1 and 2 are restricted to energy-based and cold-dissection tools (orange). 
    Training videos are non-overlapping and balanced with best effort across clients by matching video lengths across datasets. 
    The re-curated m2cai16-tool-locations subset is indicated by the red dashed box.
    VID30 and VID31 are excluded due to corrupted labels.}
    \label{fig:G_combined_distribution}
\end{figure}

We selected two public laparoscopic surgical tool detection datasets, m2cai16-tool-locations \cite{jin2018m2caiDATA} and CholecTrack20 \cite{Nwoye_2025_cholectrack20DATA}. 
While both datasets provide frame-level bounding box annotations for laparoscopic surgical tools in cholecystectomy procedures, they are curated by different institutions in the United States of America (USA) and France, respectively. 

m2cai16-tool-locations contains 10 laparoscopic surgical videos (v*) with 2811 frames in total, while CholecTrack20 includes 18 videos (VID*) with 31226 frames in total after excluding videos (VID30/31) with corrupted labels.
m2cai16-tool-locations' original train/valid/test splits suffer from temporal data leakage, as adjacent frames in the same video can appear in multiple splits.
To address this, we curated a new version of the dataset with non-overlapping videos across splits, as detailed in Figure~\ref{fig:G_combined_distribution}, circled in a red-dashed box.

\begin{figure}[t]
    \centering
    \includegraphics[width=\columnwidth]{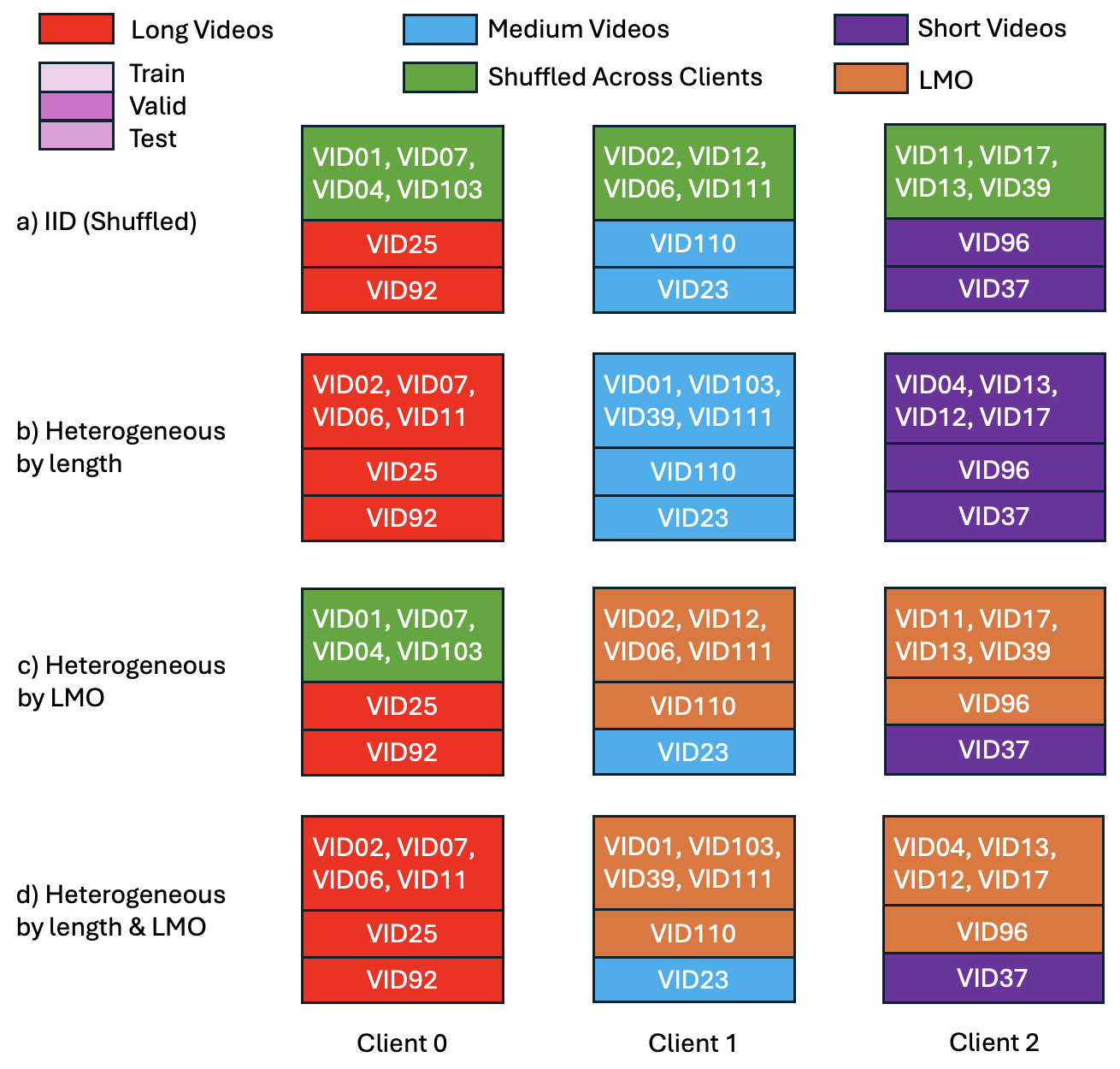}
    \caption{Dataset distribution across clients in the $G_{\text{track20}}$ experimental settings, constructed using CholecTrack20 only. 
    Within each client, cells correspond to train/validation/test splits. 
    Four settings are considered: (a) IID ($G_{\text{track20-a}}$), where training videos are pooled and randomly redistributed across clients (green); 
    (b) length-based heterogeneity ($G_{\text{track20-b}}$), where clients are assigned long, medium, or short videos; 
    (c) Leave-Multiple-Out (LMO) label heterogeneity ($G_{\text{track20-c}}$), where subsets of tool annotations are removed at specific clients (orange); 
    and (d) combined length and LMO heterogeneity ($G_{\text{track20-d}}$). 
    Training videos are non-overlapping across clients. VID30 and VID31 are excluded due to corrupted labels.}
    \label{fig:G_track20_distribution}
\end{figure}

\subsection{Experimental Design}
To address RQ1, we compare FL with per-client centralized training under In-Distribution (ID) and Cross-Distribution (CD) evaluation, where models are tested on their own and other clients’ test splits, respectively.

To investigate RQ2 and RQ3, we design two experiment groups: one using both datasets ($G_{\text{combined}}$; Fig.~\ref{fig:G_combined_distribution}) and one using only CholecTrack20 ($G_{\text{track20}}$; Fig.~\ref{fig:G_track20_distribution}).
For each group, we curate the data into three virtual institutions (clients) to study FL under controlled heterogeneity. 
This is the minimal setting that enables asymmetric client roles (e.g., fully annotated versus label-restricted sites), shows the difference between FedAvg and FedMedian, and supports cross-distribution evaluation.
Each group simulates IID and heterogeneous FL settings with these three clients, and validation and test splits are fixed across all settings in each group to ensure fair comparison.

In $G_{\text{combined}}$, we define three settings:
(a) IID ($G_{\text{combined-a}}$), where all training videos are pooled and randomly redistributed;
(b) curation-based heterogeneity ($G_{\text{combined-b}}$), where client 0 contains m2cai16-tool-locations data and clients 1–2 contain disjoint subsets of CholecTrack20; and
(c) Leave-Multiple-Out (LMO) label heterogeneity ($G_{\text{combined-c}}$), where client 0 has annotations for all tool classes, while clients 1 and 2 are restricted to energy-based and cold-dissection tools, respectively.
Training splits are constructed with non-overlapping videos and best-effort balanced frame counts across clients by matching video lengths across datasets.

Because $G_{\text{combined}}$ still exhibits residual heterogeneity due to dataset-specific curation and frame sampling, we further isolate temporal heterogeneity using $G_{\text{track20}}$, which relies solely on CholecTrack20. We define:
(a) an IID setting ($G_{\text{track20-a}}$);
(b) a video-length heterogeneous setting ($G_{\text{track20-b}}$), where clients receive sorted long, medium, or short videos; and
(c-d) two LMO label-heterogeneous settings ($G_{\text{track20-c}}$, $G_{\text{track20-d}}$), constructed analogously to $G_{\text{combined-c}}$ on top of the IID and video-length splits, respectively.

\subsection{Experimental Setup}
All experiments were conducted on a shared Graphics Processing Unit (GPU) cluster equipped with NVIDIA A100 and V100 GPUs, with runs dynamically scheduled on available resources. 
To ensure reproducibility and quantify run-to-run variability, all experiments were performed with fixed random seeds and repeated using three distinct seed sets per federated client (012, 345, and 678).

For FL, models were trained for 20 communication rounds, each consisting of 20 local epochs, with a batch size of 8. 
Before each communication round, client models were evaluated on their respective validation sets, and the best-performing client model (highest mAP50${^\text{val}}$) was selected for server-side aggregation. 
For centralized training, models were trained for 400 epochs (equivalent to 20 rounds × 20 epochs) using the same batch size, ensuring a fair comparison with FL training in terms of total optimization steps.

Across both FL and centralized settings, all remaining training hyperparameters followed the default configuration of Ultralytics YOLOv11n, including an early stopping patience of 100 epochs. 
For all FL strategies, the final aggregation step after the last communication round was omitted, as PA strategies do not yield a fully aggregated global model for deployment at individual clients.

\subsection{Evaluation Metrics}

Model performance is evaluated using mean Average Precision at an Intersection over Union (IoU) threshold of 0.5 (mAP${50}$), a standard metric for object detection \cite{Jocher_2023_Ultralytics_YOLO}. 
Performance is reported under both ID evaluation, where models are tested on the same client’s test split used for training, and CD evaluation, where models are tested on the test splits of other clients. 
Given three clients and three random seed sets per experiment, we report the mean and standard deviation of mAP${50}$ computed across all clients and seeds.

\begin{figure*}[t]
\centering
\includegraphics[width=\textwidth]{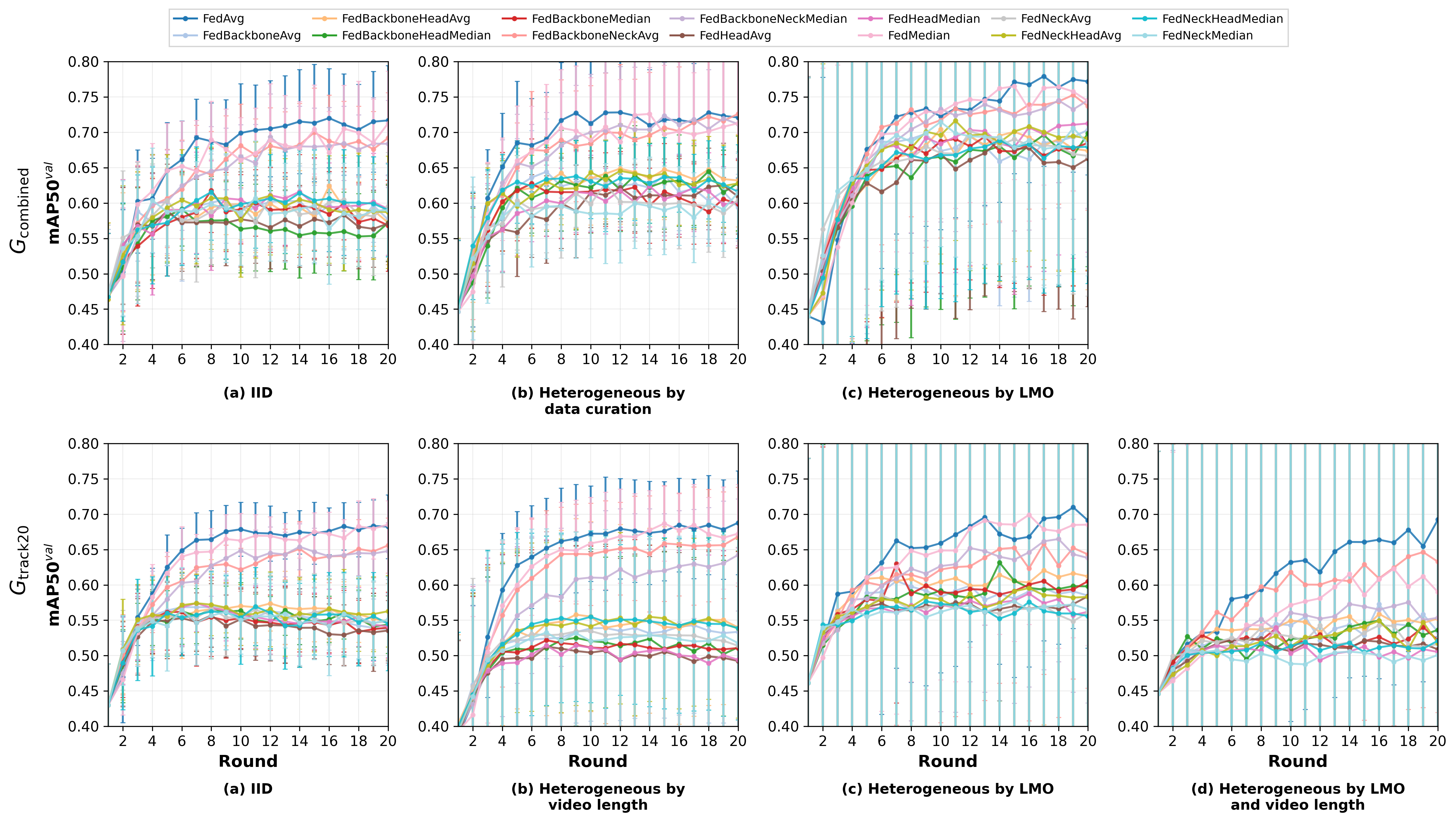}
\caption{Training progression of mAP50 under different aggregation strategies in $G_{\text{combined}}$ and $G_{\text{track20}}$ settings. First row is $G_{\text{combined}}$, and each column corresponds to a different setting: (a) IID, (b) heterogeneous by data curation, and (c) heterogeneous by LMO. 
Second row is $G_{\text{track20}}$, and each column corresponds to a different setting: (a) IID, (b) heterogeneous by video length, (c) heterogeneous by LMO, and (d) heterogeneous by LMO and video length. }
\label{fig:experiments_comparison_mAP50}
\end{figure*}

\section{Results}
In this section, we report empirical findings from federated training under different aggregation strategies and data heterogeneity settings. 
We first present results related to training progression and convergence behavior, followed by final performance evaluated on test splits.
While each experimental setting is designed to emphasize a dominant form of heterogeneityness, we note that real surgical data inherently exhibit multiple overlapping sources of heterogeneity.



\subsection{Training Progression Comparison}

Figure~\ref{fig:experiments_comparison_mAP50} shows the evolution of validation mAP50 across communication rounds for different aggregation strategies in the $G_{\text{combined}}$ and $G_{\text{track20}}$ settings.
Across all settings in both groups, FA and FedBackboneNeck consistently achieve the highest validation performance. 
These strategies maintain an upward or stable performance trend throughout the 20 communication rounds. 
Notably, FedBackboneNeck achieves comparable performance to FA at similar convergence rates while transmitting 17.73\% fewer parameters per round.
In contrast, other PA strategies typically plateau earlier in training, answering RQ2.
When comparing different settings within each group, we observe that more heterogeneous settings tend to narrow the performance gap between FA and FedBackboneNeck with other strategies, but the standard deviation also increases.

\subsection{Evaluation on Test Splits}
Final model performance evaluated on each client's test split under different settings in $G_{\text{combined}}$ is shown in Table~\ref{tab:combined_coupling_results}, while those under different settings in $G_{\text{track20}}$ are shown in Table~\ref{tab:track20_coupling_results}.
For each setting, we report the mean and standard deviation of ID and CD mAP50 across clients and random seeds. 
Centralized training results are included for comparison with top FL strategies.

We first compare FL with centralized training to answer RQ1.
Under settings in both $G_{\text{combined}}$ and $G_{\text{track20}}$, FL using standard aggregation strategies with FA or FedBackboneNeck achieves test performance comparable to centralized training. 
Despite lower standard deviations, no consistent improvement over centralized baselines' mean performance is observed for either ID or CD evaluation.
Although not shown here, the main source of standard deviation comes from variability across clients rather than random seeds, indicating that FL can narrow performance gaps among clients.

Across most settings in both experiment groups, FA and FedBackboneNeck achieve the highest ID and CD performance. 
In several cases, FedBackboneNeck matches or exceeds FA while using fewer parameters for aggregation.
This consolidates findings from training progression, indicating that these strategies can save communication costs with minimal performance trade-offs (RQ2).

Regarding RQ3, In $G_{\text{combined}}$, curation-based heterogeneity $G_{\text{combined-b}}$ yields higher ID performance than IID shuffled $G_{\text{combined-a}}$ but lower CD performance.
This is consistent with their centralized training performance.
In $G_{\text{track20}}$, heterogeneity based on video length $G_{\text{track20-b}}$ results in similar mean performance to IID $G_{\text{track20-a}}$ training but with reduced ID variance across clients.
Similarly, $G_{\text{track20-b}}$'s ID-C performance also has a low variance.
Across all LMO settings in both groups, test performance drops substantially for both ID and CD evaluation, with large standard deviations observed across clients.
Notably, in $G_{\text{track20-d}}$, FL boosted performance compared to centralized training for both ID and CD evaluation. Also, although it cannot be seen in the tables, FL does allow clients with missing labels to detect unseen items that are frequently seen by other clients, such as the grasper. This never occurred during centralized training, and it is harmful to the performance of the client with all labels.

Overall, we can observe that Table~\ref{tab:track20_coupling_results} shows higher mAP50 in all settings compared to Table~\ref{tab:combined_coupling_results}, indicating that mixing datasets with different curation protocols introduces more heterogeneity, challenging model training despite providing more data.

\begin{table*}[t]
\centering
\caption{Aggregation strategies with top 3 federated ID performance in $G_{\text{combined}}$ settings. Evaluation is performed on each client's test split. ID: in-distribution, CD: cross-distribution, and -C suffix indicates centralized training. }
\label{tab:combined_coupling_results}
\begin{tabular}{llccc|cc}
\toprule
\multirow{3}{*}{\textbf{Setting}} & \multirow{3}{*}{\textbf{Aggregation}} & \multirow{3}{*}{\textbf{\makecell{Ranking$^{\text{ID}}$\\per Setting}}} & \multicolumn{2}{c}{\textbf{Federated}} & \multicolumn{2}{c}{\textbf{Centralized}} \\
\cmidrule(lr){4-5}\cmidrule(lr){6-7}
 & & & \textbf{mAP$_{50}^{\text{ID}}$} & \textbf{mAP$_{50}^{\text{CD}}$} & \textbf{mAP$_{50}^{\text{ID-C}}$} & \textbf{mAP$_{50}^{\text{CD-C}}$} \\
 & & & \textbf{(\%)}(↑) & \textbf{(\%)}(↑) & \textbf{(\%)}(↑) & \textbf{(\%)}(↑) \\  
\midrule
\multirow{3}{*}{a) IID (Shuffled)} 
  & FedAvg & 1 & 55.7$_{\pm3.6}$ & 53.6$_{\pm5.1}$ & 53.5$_{\pm5.4}$ & 55.1$_{\pm4.3}$ \\
  & FedMedian & 2 & 54.0$_{\pm3.3}$ & 52.2$_{\pm4.8}$ &     &     \\
  & FedBackboneNeckAvg & 3 & 50.5$_{\pm6.4}$ & 49.8$_{\pm5.0}$ &     &     \\
\midrule
\multirow{3}{*}{b) Heterogeneous by Curation} 
  & FedAvg & 1 & 58.1$_{\pm4.7}$ & 50.0$_{\pm9.1}$ & 58.7$_{\pm4.9}$ & 49.8$_{\pm11.9}$ \\
  & FedMedian & 2 & 56.6$_{\pm3.3}$ & 47.1$_{\pm9.4}$ &     &     \\
  & FedBackboneNeckMedian & 3 & 55.7$_{\pm4.8}$ & 47.6$_{\pm9.5}$ &     &     \\
  \midrule
\multirow{3}{*}{c) Heterogeneous by LMO} 
  & FedMedian & 1 & 26.4$_{\pm16.3}$ & 21.6$_{\pm10.5}$ & 28.3$_{\pm19.3}$ & 23.4$_{\pm12.8}$ \\
  & FedBackboneNeckMedian & 2 & 25.1$_{\pm17.6}$ & 20.9$_{\pm8.3}$ &     &     \\
  & FedAvg & 3 & 24.4$_{\pm16.6}$ & 22.9$_{\pm10.0}$ &     &     \\
\bottomrule
\end{tabular}
\end{table*}

\begin{table*}[t]
\centering
\caption{Aggregation strategies with top 3 federated ID performance in $G_{\text{track20}}$ settings. Evaluation is performed on each client's test split. ID: in-distribution, CD: cross-distribution, and -C suffix indicates centralized training. }
\label{tab:track20_coupling_results}
\begin{tabular}{llccc|cc}
\toprule
\multirow{3}{*}{\textbf{Setting}} & \multirow{3}{*}{\textbf{Aggregation}} & \multirow{3}{*}{\textbf{\makecell{Ranking$^{\text{ID}}$\\per Setting}}} & \multicolumn{2}{c}{\textbf{Federated}} & \multicolumn{2}{c}{\textbf{Centralized}} \\
\cmidrule(lr){4-5}\cmidrule(lr){6-7}
 & & & \textbf{mAP$_{50}^{\text{ID}}$} & \textbf{mAP$_{50}^{\text{CD}}$} & \textbf{mAP$_{50}^{\text{ID-C}}$} & \textbf{mAP$_{50}^{\text{CD-C}}$} \\
 & & & \textbf{(\%)}(↑) & \textbf{(\%)}(↑) & \textbf{(\%)}(↑) & \textbf{(\%)}(↑) \\  
\midrule
\multirow{3}{*}{a) IID (Shuffled)} 
  & FedMedian & 1 & 62.3$_{\pm9.5}$ & 61.7$_{\pm5.9}$ & 65.4$_{\pm10.5}$ & 64.3$_{\pm6.3}$ \\
  & FedAvg & 2 & 61.9$_{\pm7.4}$ & 60.6$_{\pm6.3}$ &     &     \\
  & FedBackboneNeckAvg & 3 & 61.7$_{\pm9.0}$ & 61.0$_{\pm7.3}$ &     &     \\
\midrule
\multirow{3}{*}{b) Heterogeneous by length} 
  & FedAvg & 1 & 61.5$_{\pm2.6}$ & 62.4$_{\pm6.9}$ & 61.6$_{\pm1.7}$ & 62.0$_{\pm8.3}$ \\
  & FedMedian & 2 & 61.1$_{\pm2.3}$ & 62.6$_{\pm7.4}$ &     &     \\
  & FedBackboneNeckAvg & 3 & 60.9$_{\pm2.9}$ & 61.5$_{\pm6.5}$ &     &     \\
  \midrule
\multirow{3}{*}{c) Heterogeneous by LMO} 
  & FedAvg & 1 & 34.2$_{\pm17.6}$ & 33.8$_{\pm16.4}$ & 34.3$_{\pm22.0}$ & 34.8$_{\pm21.8}$ \\
  & FedMedian & 2 & 34.1$_{\pm19.6}$ & 31.7$_{\pm16.5}$ &     &     \\
  & FedBackboneNeckAvg & 3 & 33.7$_{\pm20.0}$ & 31.7$_{\pm17.3}$ &     &     \\
    \midrule
\multirow{3}{*}{d) Heterogeneous by Length \& LMO} 
  & FedAvg & 1 & 35.9$_{\pm19.5}$ & 35.6$_{\pm17.8}$ & 29.6$_{\pm24.9}$ & 32.0$_{\pm25.2}$ \\
  & FedBackboneNeckAvg & 2 & 31.4$_{\pm22.5}$ & 30.5$_{\pm20.8}$ &     &     \\
  & FedNeckHeadMedian & 3 & 29.5$_{\pm25.3}$ & 28.0$_{\pm20.2}$ &     &     \\
\bottomrule
\end{tabular}
\end{table*}

\section{Discussion}
Our work introduces the first FL framework for surgical object detection. It is also the first to investigate YOLO's behavior under FA and PA with IID and heterogeneous surgical data. We structure the discussion according to the three RQs posed in Section~\ref{sec:intro}.

\subsection{RQ1: FL versus Centralized Single-Institution Training}

Our results show that standard FL aggregators show no consistent improvement over centralized baselines' mean performance. This is expected as only common FL aggregators are used without any SOTA personalization techniques.
However, FL can narrow performance gaps among clients, as indicated by lower standard deviations compared to centralized training.
This suggests that FL can help institutions with limited resources benefit from knowledge sharing with larger centers, leading to more equitable model performance across institutions.

\subsection{RQ2: Effects of PA Strategies on Performance-Communication Trade-offs}

Regarding RQ2, FedBackboneNeck strategies consistently outperform other PA strategies across different settings, achieving comparable and sometimes even better performance than FA while reducing overall communication costs.
This suggests that both the backbone and neck components of YOLO play a more critical role in shared feature extraction and representation learning, while the head component is more specialized and can be trained locally without sacrificing performance greatly.

\subsection{RQ3: Impact of Clinically Motivated Data Heterogeneity on Federated YOLO Performance}

Moving on to RQ3, our results demonstrate that not all heterogeneity is equally detrimental; rather, its impact depends on how it shapes and balances intra-client consistency and inter-client divergence.

In the combined-dataset setting, the curation-based heterogeneous configuration ($G_{\text{combined-b}}$) exhibits superior ID performance compared to the more IID shuffled setting ($G_{\text{combined-a}}$), despite being more heterogeneous overall.
This counterintuitive result can be attributed to increased internal consistency within each client, as each client contains data from a single dataset, which facilitates local convergence and enables the model to capture dataset-specific visual characteristics more effectively.
However, this benefit comes at the cost of reduced CD generalization, as evidenced by poorer CD and CD-C performance.
These results indicate a trade-off: curation-based heterogeneity can enhance ID performance while simultaneously hindering generalization across distributions, an effect that FL can partially mitigate through collaborative training.

In the video-length-based heterogeneous setting ($G_{\text{track20-b}}$), although the mean ID performance is slightly lower than in the IID case ($G_{\text{track20-a}}$), performance variance across clients is substantially reduced.
This suggests that grouping videos by length imposes a form of structural consistency related to procedural duration and workflow, leading to more stable learning dynamics across clients and more consistent federated outcomes.

In contrast, LMO heterogeneity has a uniformly detrimental effect.
Across all LMO settings ($G_{\text{combined-c}}$, $G_{\text{track20-c}}$, and $G_{\text{track20-d}}$), both ID and CD performance degrade markedly.
Missing annotations for specific classes bias feature representations and impair the model’s ability to learn discriminative and transferable features.
Standard FL aggregators are insufficient to address this challenge, underscoring the need for more advanced strategies tailored to severe label imbalance and absence.
Nevertheless, in $G_{\text{track20-d}}$, FL still improves both ID and CD performance relative to centralized training, reinforcing the observation that increased intra-client consistency can render otherwise harmful heterogeneity more tractable for FL.

Overall, these findings demonstrate that heterogeneity should not be treated as a monolithic challenge.
Heterogeneity that improves internal client consistency—such as curation- or duration-based partitioning—can be beneficial under FL, whereas missing label-based heterogeneity poses a fundamental limitation that requires dedicated methodological solutions.

\subsection{Limitations and Future Work}

As the first FL framework tailored to surgical object detection, UltraFlwr has several limitations that motivate future work.
First, while YOLO is a strong and widely adopted baseline, our study is limited to a single architectural family.
Extending UltraFlwr to alternative detectors such as Faster R-CNN \cite{ren2015fasterRCNN} would help assess the generalizability of our findings across different model architectures.

Second, our evaluation is primarily conducted in simulated FL settings.
Although this enables controlled experimentation, it does not fully capture the constraints of real-world clinical deployment.
Future work should therefore focus on live or near-live surgical settings to evaluate robustness, latency, and reliability under realistic operating conditions.
Preliminary lab-based tests are performed by us on edge devices, including Raspberry Pi and NVIDIA Jetson platforms, suggesting feasibility, but systematic evaluation on heterogeneous hardware—particularly in resource-constrained settings such as Low- and Middle-Income Countries (LMICs)—remains necessary.

Finally, while we investigate several common forms of heterogeneity in laparoscopic cholecystectomy datasets, our analysis does not capture the full range of clinically relevant variability.
Observable distributional differences often reflect deeper underlying factors, including patient characteristics, pathology, surgeon expertise, institutional protocols, and acquisition hardware.
Future progress would benefit from federated surgical datasets enriched with detailed yet anonymized metadata to enable more clinically grounded heterogeneity modeling and the development of more robust FL methods.

\section{Conclusion}
This work provides a systematic analysis of federated YOLO training under full and partial aggregation across clinically motivated heterogeneous settings, offering practical insights for federated surgical object detection.
We demonstrate that PA strategies, particularly FedBackboneNeck, can reduce communication overhead while maintaining performance comparable to FA.
Our results further show that the impact of heterogeneity depends strongly on intra-client consistency, with curation- and duration-based partitioning yielding more stable federated outcomes than label-missing scenarios.
Overall, UltraFlwr establishes a flexible foundation for privacy-preserving, multi-institutional collaboration and advances the deployment of FL in surgical computer vision.


\bibliographystyle{ieeetran}
\bibliography{ref}

\end{document}